\documentclass[11pt,a4paper]{article}
\usepackage[hyperref]{emnlp2018}
\usepackage{times}
\usepackage{latexsym}
\usepackage{amsmath}
\usepackage{amsmath}	% math fonts
\usepackage{amsthm}
\usepackage{amsfonts}
\usepackage{amssymb}
\usepackage{url}
\usepackage{adjustbox}
\usepackage{color}
\usepackage{xspace}
\usepackage{relsize}
\usepackage{multirow}
\usepackage{float}
\usepackage{multirow}
\usepackage{tkz-euclide}
\usetkzobj{all}
\usetikzlibrary{shadings}
\usepackage{cleveref}

\usepackage{subcaption}
\usepackage{nth}
\usepackage{wrapfig}
\usepackage{tikz,pgfplots}
\pgfplotsset{compat=newest}
\usetikzlibrary{plotmarks, shapes, arrows, positioning, shadows, trees}

\newlength\figureheight 
	\newlength\figurewidth

\newcommand\BLEU{\textsc{Bleu}\xspace}
\newcommand\TER{\textsc{Ter}\xspace}
\newcommand\PPL{\textsc{Ppl}\xspace}

\newcommand{\R}{\mathbb{R}}
\usepackage{url}

\aclfinalcopy % Uncomment this line for the final submission
\def\aclpaperid{***} %  Enter the acl Paper ID here

\title{Towards Two-Dimensional Sequence to Sequence Model in Neural Machine Translation}

 \author{Parnia Bahar, Christopher Brix and Hermann Ney\\
    Human Language Technology and Pattern Recognition Group \\
    Computer Science Department \\
    RWTH Aachen University \\
    D-52056 Aachen, Germany \\
    {\tt <surname>@cs.rwth-aachen.de }}

\date{}

\begin{document}
\maketitle
\begin{abstract}

This work investigates an alternative model for neural machine translation (NMT) and proposes a novel architecture, where we employ a multi-dimensional long short-term memory (MDLSTM) for translation modeling. In the \mbox{state-of-the-art} methods, source and target sentences are treated as one-dimensional sequences over time, while we view translation as a two-dimensional (2D) mapping using an \mbox{MDLSTM} layer to define the correspondence between source and target words. We extend beyond the current sequence to sequence backbone NMT models to a 2D structure in which the source and target sentences are aligned with each other in a 2D grid.
Our proposed topology shows consistent improvements over attention-based sequence to sequence model on two WMT 2017 tasks, German$\leftrightarrow$English.
\end{abstract}

\section{Introduction}

The widely used state-of-the-art neural \mbox{machine} translation (NMT) systems are based on an encoder-decoder architecture equipped with attention layer(s). The encoder and the decoder can be constructed using \mbox{recurrent} neural networks (RNNs), especially long-short term memory (LSTM) \cite{Bahdanau_15_attention, GNMT}, convolutional neural networks (CNNs) \cite{conv_seq2seq}, self-attention units \cite{transformer}, or a combination of them \cite{best_of_2}.
In all these architectures, source and target sentences are handled separately as a one-dimensional sequence over time. Then, an \mbox{attention} mechanism (additive, multiplicative or multihead) is incorporated into the decoder to selectively focus on individual parts of the source sentence. 

One of the weaknesses of such models is that the encoder states are computed only once at the beginning and are left untouched with \mbox{respect} to the target histories. In this case, at every decoding step, the same set of \mbox{vectors} are read repeatedly. Hence, the attention mechanism is limited in its ability to \mbox{effectively} model the coverage of the source sentence.
By providing the encoder states with the greater capacity to remember what has been generated and what needs to be translated, we believe that we can alleviate the coverage problems such as over- and under-translation. 
% Moreover it is similar to the way humans do translate. Human translators simultaneously take the complete target history and the whole source sequence into account. 

One solution is to assimilate the context from both source and target sentences jointly and to align them in a two-dimensional grid. Two-dimensional LSTM (2DLSTM) is able to process data with complex interdependencies in a 2D space \cite{Graves_12_book}. 
% it therefore seems attractive and inspiring to apply the advantage of it in NMT. 

To incorporate the solution, in this work, we propose a novel architecture based on the \mbox{2DLSTM} unit, which enables the computation of the encoding of the source sentence as a function of the previously generated target words.
We treat translation as a 2D mapping. One dimension processes the source sentence, and the other dimension generates the target words. Each time a target word is generated, its representation is used to compute a hidden state sequence that models the source sentence encoding. 
In principle, by updating the encoder states across the second dimension using the target history, the 2DLSTM captures the coverage concepts internally by its cell states. 

\section{Related Works}
MDLSTM \cite{Graves_08_thesis, Graves_12_book} has been successfully used in handwriting recognition (HWR) to automatically extract features from raw images which are inherently two-dimensional \cite{Graves_08_md_hwr,Leifert_16_md_hwr, Voigtlaender_16_hwr}.   
\newcite{Voigtlaender_16_hwr} explore a larger MDLSTM for deeper and wider architectures using an implementation for the graphical processing unit (GPU).
It has also been applied to automatic speech recognition (ASR) where a 2DLSTM scans the input over both time and frequency jointly \cite{Li_16_md_asr, Sainat_16_md_asr}.
As an alternative architecture to the concept of \mbox{MDLSTM}, \newcite{Kalchbrenner_15_grid_lstm} propose a grid LSTM that is a network of LSTM cells arranged in a multidimensional grid, in which the cells are communicating between layers as well as time recurrences.
\newcite{Li_17_google_grid_lstm} also apply the grid LSTM architecture for the endpoint detection task in ASR. 

This work, for the first time, presents an end-to-end 2D neural model where we process the source and the target words jointly by a 2DLSTM layer.

\section{Two-Dimensional LSTM} \label{sec_2dlstm}

\begin{figure}[h]
\centering
\includegraphics[width=0.49\textwidth]{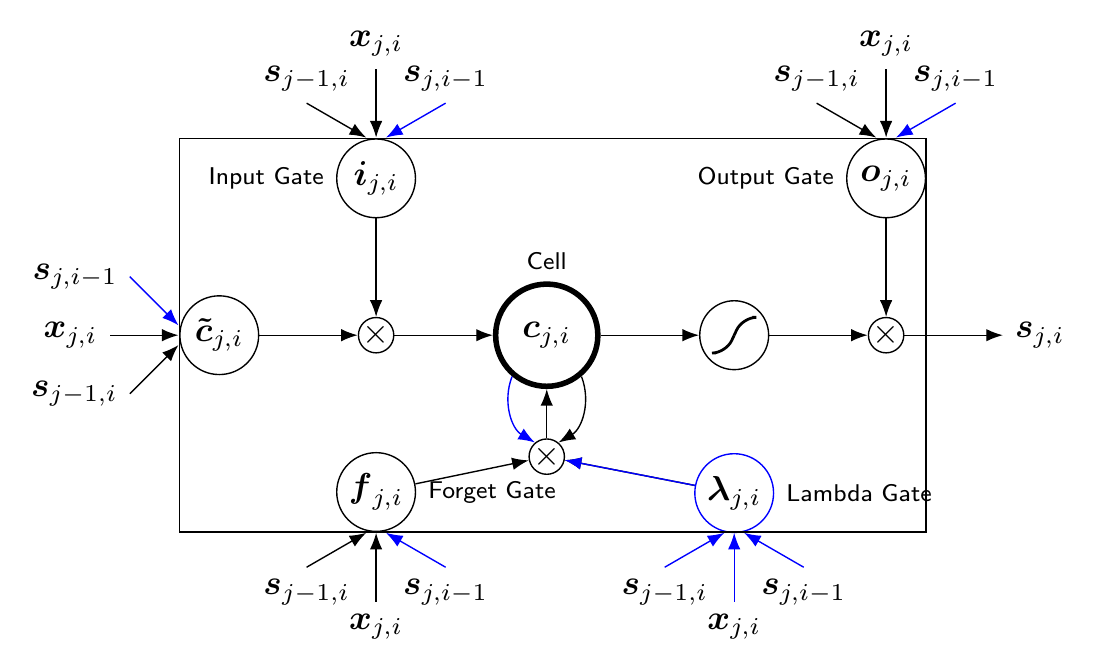}
\caption{2DLSTM unit. The additional links vs. standard LSTM are marked in blue.} \label{fig:2dlstm}
\end{figure}

The 2DLSTM has been introduced by \cite{Graves_08_thesis} as a generalization of standard LSTM. Figure \ref{fig:2dlstm} illustrates one of the stable variants proposed by \cite{stable_cells}.
A 2DLSTM unit processes a 2D sequential data %$\boldsymbol{x}_{1,1}^{J,I}$ 
$x \in \R^{J \times I}$ 
of arbitrary lengths, $J$ and $I$.
At time step $(j,i)$, the computation of its cell depends on both vertical ${s}_{j, i-1}$ and horizontal hidden states ${s}_{j-1, i}$ (see \Crefrange{mdlstm:1}{mdlstm:lambda}).
Similar to the LSTM cell, it maintains some state information in an internal cell state ${c}_{j,i}$.
Besides the input ${i}_{j,i}$, the forget ${f}_{j,i}$ and the output ${o}_{j,i}$ gates that all control information flows, 2DLSTM employs an extra \texttt{lambda gate} ${\lambda}_{j,i}$. As written in Equ. \ref{mdlstm:lambda}, its activation is computed analogously to the other gates. 
The lambda gate is used to weight the two predecessor cells ${c}_{j-1,i}$ and ${c}_{j,i-1}$ before passing them through the forget gate (Equation \ref{eq:mdlstm:c}). 
%The remaining computations are analogue to LSTM. 
$g$ and $\sigma$ are the $\tanh$ and the sigmoid functions. $V$s, $W$s and $U$s are the weight matrices.

In order to train a 2DLSTM unit, back-propagation through time (BPTT) is performed over two dimensions \cite{Graves_08_thesis, Graves_12_book}.   
Thus, the gradient is passed backwards from the time step $(J,I)$ to $(1,1)$, the origin.
More details, as well as the derivations of the gradients, can be found in \cite{Graves_08_thesis}. 
\begin{align}
{i}_{j,i} &= \sigma \Big( W_{1}{x}_{j,i} + U_{1}{s}_{j-1, i} + V_{1}{s}_{j, i-1}   \Big) \label{mdlstm:1}\\ 
{f}_{j,i} &= \sigma \Big( W_{2}{x}_{j,i} + U_{2}{s}_{j-1, i} + V_{2}{s}_{j, i-1} \Big) \\ 
{o}_{j,i} &= \sigma \Big( W_{3}{x}_{j,i} + U_{3}{s}_{j-1, i} + V_{3}{s}_{j, i-1}  \Big)  \\ 
{\tilde{c}}_{j,i} &= g \Big( W_{4}{x}_{j,i} + U_{4}{s}_{j-1, i} + V_{4}{s}_{j, i-1}  \Big) \\
\label{mdlstm:lambda} 
{\lambda}_{j,i} &= \sigma \Big( W_{5}{x}_{j,i} + U_{5}{s}_{j-1, i} + V_{5}{s}_{j, i-1}  \Big) \\
\label{eq:mdlstm:c}
{c}_{j,i} &=  {f}_{j,i}  \circ \big[ {\lambda}_{j,i}  \circ {c}_{j-1,i} + (1- {\lambda}_{j,i}) \circ {c}_{j,i-1} \big] \nonumber \\
&\phantom{==} +  {\tilde{c}}_{j,i} \circ {i}_{j,i}   \\
{s}_{j,i} &=
g \left( {c}_{j,i} \right) \circ {o}_{j,i}
\end{align}

\section{Two-Dimensional Sequence to Sequence Model} \label{2d-seq-to-seq}

\begin{figure}[h]
  \centering
\includegraphics[width=0.46\textwidth]{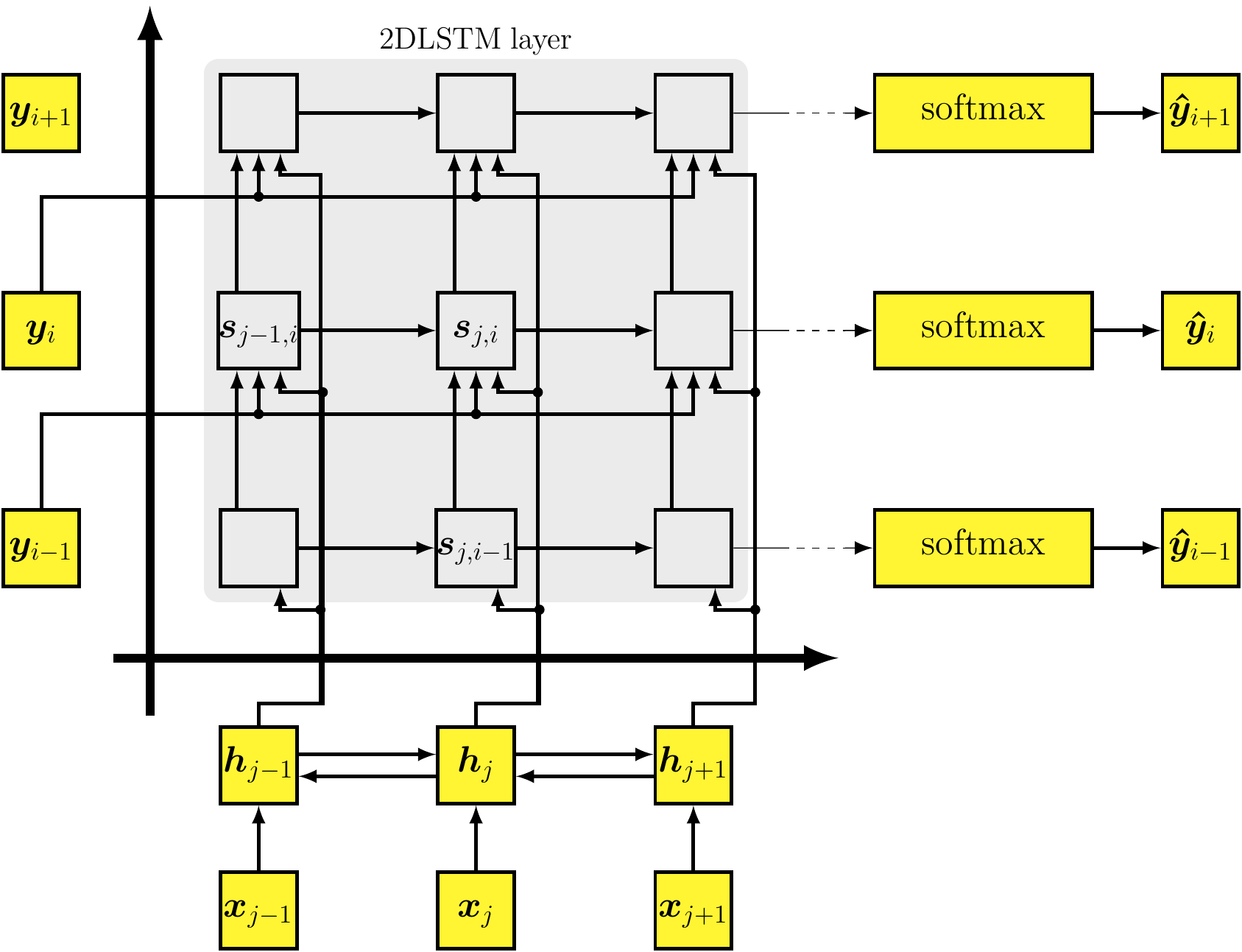}
\caption{Two-dimensional sequence to sequence model (2D-seq2seq).} \label{fig:2dlstm_seq2seq}
\end{figure}

We aim to apply a 2DLSTM to map the source and the target sequences into a 2D space as shown in Figure \ref{fig:2dlstm_seq2seq}. We call this architecture, the two-dimensional sequence to sequence (\texttt{2D-seq2seq}) model. 

Given a source sequence $x_1^{J}= x_1,\ldots, x_{J}$ and a target sequence $y_1^{I}= y_1,\ldots, y_{I}$, we scan the source sequence from left to right and the target sequence from bottom to top as shown in Figure \ref{fig:2dlstm_seq2seq}.
In the 2D-seq2seq model, one dimension of the \mbox{2DLSTM} (horizontal-axis in the figure) serves as the encoder and another (vertical axis) plays the role of the decoder. 
As a pre-step before the 2DLSTM, in order to have the whole source context, a bidirectional LSTM scans the input words once from left to right and once from right to left to compute a sequence of encoder states ${h}_{1}^{J}= {h}_{1},\ldots, {h}_{J}$. 
At time step $(j,i)$, the 2DLSTM receives both encoder state, ${h}_{j}$, and the last target embedding vector, ${y}_{i-1}$, as an input. 
It repeatedly updates the source information, ${h}_{1}^{J}$, while generating new target word, ${y}_{i}$. 
The state of the 2DLSTM is computed as follows. 
% For simplicity, we betake the weight matrices. 
% We note that $\boldsymbol{f}_{j}$ and $\boldsymbol{e}_{i-1}$ refer to the embedded vectors.
\begin{align}
{s}_{j,i} = \psi \Big( W \big[{h}_{j}; {y}_{i-1}  \big], U{s}_{j-1,i}, V{s}_{j,i-1}\Big)
\end{align}
where $\psi$ stands for the 2DLSTM as a function. At each decoder step, once the whole source sequence is processed from $1$ to $J$, the last hidden state of the 2DLSTM, ${s}_{J,i}$, is used as the context vector. It means, at time step $i$, ${t}_{i} = {s}_{J,i}$. 
In order to generate the next target word, ${y}_{i}$, a transformation followed by a softmax operation is applied. Therefore:
\begin{align}
p_i(y_i = w | y_1^{i-1}, x_1^{J})  = \frac{ \exp( W_o {t}_{iw})} {\sum_{v=1}^{|V_t|} \exp(W_o {t}_{iv})}
\end{align}
where $W_o$ and $|V_t|$ are the weight matrix and the target vocabulary respectively. 

\subsection{Training versus Decoding}
One practical concern that should be noticed is the difference between the training and the decoding. 
Since the whole target sequence is known during training, all states of the \mbox{2DLSTM} can be computed once at the beginning.
Slices of it can then be used during the forward and backward training passes. In theory, the complexity of training is $\mathcal{O}(JI)$. 
But, in practice, the training computation can be optimally parallelized to take linear time \cite{Voigtlaender_16_hwr}.
% In contrast to the training, 
During the decoding, only the already generated target words are available. Thus, either all 2DLSTM states have to be recomputed, or it has to be extended by an additional row at every time step $i$ that cause higher complexity.

\begin{table*}
\centering
\begin{adjustbox} {max width=0.64\textheight}  
\begin{tabular}{|c|l|c|c|c|c|c|c|c|c|c|c|c|}
\hline
\multicolumn{2}{|c|}{\multirow{3}{*}{Models}} & \multirow{3}{*}{Hidden Size} & \multicolumn{5}{c|}{De$\to$En}   & \multicolumn{5}{c|}{En$\to$De}  \\ \cline{4-13} 
\multicolumn{2}{|c|}{} & & \bf{devset} & \multicolumn{2}{l|}{\bf{newstest2016}}     & \multicolumn{2}{c|}{\bf{newstest2017}}  & \bf{devset}  & \multicolumn{2}{c|}{\bf{newstest2016}}    & \multicolumn{2}{l|}{\bf{newstest2017}}   \\ \cline{4-13} 
\multicolumn{2}{|c|}{} & & \PPL & \multicolumn{1}{c|}{\BLEU} & \multicolumn{1}{c|}{\TER} & \multicolumn{1}{c|}{\BLEU} & \multicolumn{1}{c|}{\TER} & \PPL & \multicolumn{1}{c|}{\BLEU} & \multicolumn{1}{c|}{\TER} & \multicolumn{1}{c|}{\BLEU} & \multicolumn{1}{c|}{\TER} \\ \hline \hline

1  & attention &  \multirow{3}{*}{n=500} & 7.3 & 31.9 & 48.6 & 27.5 & 53.1 & 7.0& 27.0& 53.9 & 22.1 &  60.5 \\

2  &  2D-seq2seq &  & 6.5 & \bf{32.6}& 47.8 & \bf{28.2} & 52.7  & 6.1 & \bf{27.5} & 53.8 & \bf{22.4} &  60.6 \\

3  &  + weighting &  & 6.5& 32.3 & \bf{47.1} & 27.9 & \bf{51.7}  & 6.3&  \bf{27.5}& \bf{53.3} & \bf{22.4} & \bf{60.0}  \\ \hline

1  & attention & \multirow{3}{*}{n=1000} & 6.4 & 33.1 & 47.5 &  29.0&  \bf{51.9}& 6.5 &27.4  & 53.9 & 22.9 & 60.2  \\

2  &  2D-seq2seq &  & 5.7 & \bf{33.7}& \bf{46.9} & \bf{29.3} & \bf{51.9}  & 5.3& \bf{28.9} &  \bf{52.6}&  \bf{23.2}&  \bf{59.5} \\

3  &  + weighting & & 6.1 & 32.7 & 47.1 & 28.0 & \bf{51.9}  & 5.7& 27.8 & 53.0 & 22.7 &  60.0 \\ \hline \hline

4  & coverage  &  \multirow{2}{*}{n=1000} & 6.3 & 33.1 &47.5  & 28.7 & 51.9 & 5.8 & 28.6 & 52.4 & 23.0 &  59.4 \\  

5  & fertility &  \multirow{3}{*}{} & 6.2 & 33.4 & 46.9 & 28.9 & 51.6 & 5.8& 28.4& 52.1& 23.2 & 59.1   \\ \hline

\end{tabular}
\end{adjustbox}
\caption{\BLEU[\%] and \TER[\%] on the test sets and perplexity (\PPL) on the development set.}
\label{results_all_together}
\end{table*}

\section{Experiments} \label{exp}

We have done the experiments on the WMT 2017 German$\to$English and English$\to$German news tasks consisting of $4.6$M training samples collected from the well-known data sets \texttt{Europarl-v7}, \texttt{News-Commentary-v10} and \texttt{Common-Crawl}. 
We use \texttt{newstest2015} as our development set and \texttt{newstest2016} and -\texttt{2017} as our test sets, 
which contain $2169$, $2999$ and $3004$ sentences respectively. 
% We emphasize that we apply no synthetic data, no data filtering, no additional features like a language model and the ensemble of deep networks. 
No synthetic data and no additional features are used.
Our goal is to keep the baseline model simple and standard to compare methods rather that advancing the state-of-the-art systems. 

After tokenization and true-casing using \texttt{Moses} toolkit \cite{moses}, byte pair encoding (BPE) \cite{Sennrich2016:bpe} is used jointly with $20$k merge operations. We remove sentences longer than $50$ subwords and batch them together with a batch size of $50$. All models are trained from scratch by the Adam optimizer \cite{adam}, dropout of $30\%$ \cite{dropout} and the norm of the gradient is clipped with the threshold of $1$. The final models are the average of the $4$ best checkpoints of a single run based on the perplexity on the development set  \cite{averaging}. Decoding is performed using beam search of size $12$, without ensemble of various networks.

We have used our in-house implementation of the NMT system which relies on Theano
% \footnote{http://deeplearning.net/software/theano/} 
\cite{theano} and Blocks
% \footnote{https://github.com/mila-udem/blocks-examples}
\cite{blocks}. Our implementation of 2DLSTM is based on CUDA code adapted from \cite{Voigtlaender_16_hwr, Returnn2018}, leveraging some speedup. 
% considerations and the ability to use GPUs.

The models are evaluated using case-sensitive \BLEU \cite{bleu} computed by \texttt{mteval-v13a}\footnote{ftp://jaguar.ncsl.nist.gov/mt/resources/mteval-v13a.pl} 
and case-sensitive \TER \cite{ter} using \texttt{tercom}\footnote{http://www.cs.umd.edu/~snover/tercom/}. 
We also report perplexities on the development set.

\textbf{Attention Model:}
the attention based sequence to sequence model \cite{Bahdanau_15_attention} is selected as our baseline that performs quite well. 
The model consists of one layer bidirectional encoder and a unidirectional decoder with an additive attention mechanism. 
All words are projected into a $500$-dimensional embedding on both sides. To explore the performance of the models with respect to hidden size, we try LSTMs \cite{Hochreiter:1997:LSTM} with both  $500$ and $1000$ nodes. 
% The attention model is trained with the learning rate of $0.001$.

\textbf{2D-Seq2Seq Model:} 
we apply the same embedding size of that of the attention model. The 2DLSTM, as well as the bidirectional LSTM layer, are structured using the same number of nodes ($500$ or $1000$). The 2D-seq2seq model is trained with the learning rate of $0.0005$ vs. $0.001$ for the attention model.

% \subsection{Results}
\textbf{Translation Performance:} in the first set of experiments, we compare the 2D-seq2seq model with the attention sequence to sequence model. The results are shown in Table \ref{results_all_together} in the rows $1$ and $2$. 
As it is seen, for size $n=500$, the 2D-seq2seq model outperforms the standard attention model on average by $0.7\%$ \BLEU and $0.6\%$ \TER on De$\to$En, $0.4\%$ \BLEU and no improvements in \TER on En$\to$De. 
The model is also superior for larger hidden size ($n=1000$) on average by $0.5\%$ \BLEU and $0.3\%$ \TER on De$\to$En, $0.9\%$ \BLEU and $1.0\%$ \TER on En$\to$De. In both cases, the perplexity of the 2D-seq2seq model is lower compared to that of the attention model.

The 2D-seq2seq topology is analogous to the bidirectional encoder-decoder model without attention. To examine whether the 2DLSTM reduces the need of attention, in the second set of experiments, we equip our model with a weighted sum of 2DLSTM states, ${t}_{i}$, over $j$ positions  to dynamically select the most relevant information. In other words:
\begin{align}
\gamma_{j,i} &= \underset{j}{softmax} \Big( {v}^T \tanh \big( W {s}_{j,i} \big) \Big) \label{energy} \\
{t}_{i} &= \sum_{j=1}^{J} \gamma_{j,i} {s}_{j,i}  
\end{align}
In these equations, $\gamma_{j,i}$ is the normalized weight over source positions, ${s}_{j,i}$ is the \mbox{2DLSTM} states and $W$ and $v$ are weight matrices.
% It is assumed that weighting tells which source word(s) to focus on, while information about the coverage of the source remains encoded in the hidden states. 
As the results shown in the Table \ref{results_all_together} in the rows $2$ and $3$, adding an additional weighting layer on top of the 2DLSTM layer does not help in terms of \BLEU and rarely helps in \TER.

By updating the encoder states across the second dimension with respect to the target history, the 2D-seq2seq model can internally indicate which source words have already been translated and where it should focus next. Therefore, it reduces the risk of over- and under-translation. To examine our assumption, we compare the 2D-seq2seq model with two NMT models where the concepts such as fertility and coverage have been addressed \cite{Tu_coverage_16,struc_bias_Cohn_16}.

\begin{table*}
\centering
\begin{adjustbox} {max width=0.65\textheight}  
\begin{tabular}{p{1.9cm}|p{15.9cm}}
\hline
source    &  HP besch{\"a}ftigte zum Ende des Gesch{\"a}ftsjahres 2013/14 noch rund 302.000 Mitarbeiter. \\ \hline

reference & At the end of the 2013/14 business year HP still employed around 302,000 staff. \\ \hline
attention & \textcolor{red}{At the end of the financial year}, HP employed some 302,000 employees \textcolor{red}{at the end of the financial year} of 2013/14.  \\ \hline
2D-seq2seq &   HP still employs about 302,000 people at the end of the financial year 2013/14. \\ \hline
coverage & HP employed around 302,000 employees at the end of the fiscal year 2013/14. \\ \hline
fertility & HP employed some 302,000 people at the end of the fiscal year 2013/14. \\ \hline
\end{tabular}
\end{adjustbox}
\caption{An example of over-translation.}
\label{table:undertranslation}
\end{table*}

\textbf{Coverage Model:}
in the coverage model, we feed back the last alignments from the time step $i-1$ to compute the attention weight at time step $i$. Therefore, in the coverage model, we redefine the attention weight, $\alpha_{i,j}$, as:

\begin{equation}
\alpha_{i,j} = a \big(s_{i-1}, h_j, {\alpha}_{i-1,j} \big)
\label{equ:coverage}
\end{equation}
where $a$ is an attention function followed by the softmax. $h_j$ and $s_{i-1}$ are the the encoder and the previous decoder states respectively. In our experiments, we use additive attention similar to \cite{Bahdanau_15_attention}.

\textbf{Fertility Model:}
in the fertility model, we feed back the sum of the alignments over the past decoder steps to indicate how much attention has been given to the source position $j$ up to step $i$ and divide it over the fertility of source word at position $j$. This term depends on the encoder states and it varies if the word is used in a different context \cite{Tu_coverage_16}.

\begin{equation}
\beta_{i,j} = \frac{1}{N \cdot \sigma ({\upsilon}^\top_{\phi} \cdot {h_j})}\sum_{k=1}^{i-1} \alpha_{i,j}
\label{equ:with_fertility} 
\end{equation}

\begin{equation}
\alpha_{i,j} = a \big(s_{i-1},h_j, {\beta}_{i,j} \big)
\label{equ:energy_no_fertility}
\end{equation}
% where $\phi_j$ refers to the fertility of source word at position $j$. This term depends on the encoder states, because it can vary if the word is used in a different context. Like~\cite{Tu_coverage_16} in our model $\phi_j$ is defined as:
% \begin{equation}
% \phi_j = N \cdot \sigma ({\upsilon}^\top_{\phi} \cdot {h_j})
% \label{phi} 
% \end{equation}
where $N$ specifies the maximum value for the fertility which set to~2 in our experiments. $\upsilon_{\phi}$ is a weight vector.

As it is seen in Table \ref{results_all_together}, rows $2$, $4$ and $5$, our proposed model is $0.3\%$ \BLEU ahead and $0.3\%$ \TER worse compared to the fertility approach and slightly better compared to the coverage one. 
We note, the fertility and coverage models were trained using embedding size of  $620$.

We have also qualitatively verified the coverage issue in Table \ref{table:undertranslation} by showing an example from the test set.
Without the knowledge of which source words have already been translated, the attention layer is at risk of attending to the same positions multiple times.
This could lead to over-translation.
Similarly, under-translation could be occur when the attention model rarely focusing at the corresponding source positions.
As shown in the example, the \mbox{2DLSTM} can internally track which source positions have already contributed to the target generation.

\textbf{Speed:} we have also compared the models in terms of speed on a single GPU training.  
In general, the training and decoding speed of the 2D-seq2seq model is $791$ and $0.7$ words/s respectively compared to those of standard attention model which is $2944$ and $48$ words/s. 
The computation of the added weighting mechanism is negligible in this case. 
This is still an initial architecture which indicates the necessity of multi-GPU usage. We also expect to speedup the decoding phase by avoiding the unnecessary recomputation of previous \mbox{2DLSTM} states. In the current implementation, at each target step, we re-compute the 2DLSTM states from time step $0$ to $i-1$, while we only need to store the states from the last step $i-1$. This does not influence our results, as it is purely an implementation issue, not algorithm. However, decoding will still be slower than the training. One suggestion for further speedup of training phase is applying truncated BPTT on  both directions to reduce the number of updates.

The 2DLSTM can be simply combined with self-attention layers \cite{transformer} in the encoder and the decoder for better context representation as well as RNMT+ \cite{ best_of_2} that is composed of standard LSTMs. 
We believe that 2D-seq2seq model can be potentially applied to the other applications where sequence to sequence modeling is helpful. 
% In sequence modeling, classes are discrete label sequences (e.g. words, characters, phonemes, generalized triphone states, POS taggers, etc.), whose length S usually is not known ${c_1,...,c_S}$. Given an input represented by a sequence of observations ${x_1 ,...,x_T}$ with length T, statistical sequence classification based on Bayes decision rule requires a model for the label sequence posterior probability distribution $p(c_1, \cdots, c_S | x_1 , \cdots, x_T)$. This conditional distribution might be modeled using \mbox{2DLSTM} provided with empirical experiments and analysis the results.

\section{Conclusion and Future Works}
We have introduced a novel 2D sequence to sequence model (2D-seq2seq), a network that applies a 2DLSTM unit to read both the source and the target sentences jointly. 
% In the proposed method, one dimension processes the source sentence and the other generates the target sentence. 
Hence, in each decoding step, the network implicitly updates the source representation conditioned on the generated target words so far. 
The experimental results show that we outperform the attention model on two WMT 2017 translation tasks. We have also shown that our model implicitly handles the coverage issue.

As future work, we aim to develop a bidirectional \mbox{2DLSTM} and consider stacking up \mbox{2DLSTMs} for a deeper model.
We consider the results promising and try more language pairs and fine-tune the hyperparameters.

\section*{Acknowledgements}
\begin{wrapfigure}{l}{0.19\textwidth}
\vspace{-4mm}
    \begin{center}
        \includegraphics[width=0.18\textwidth]{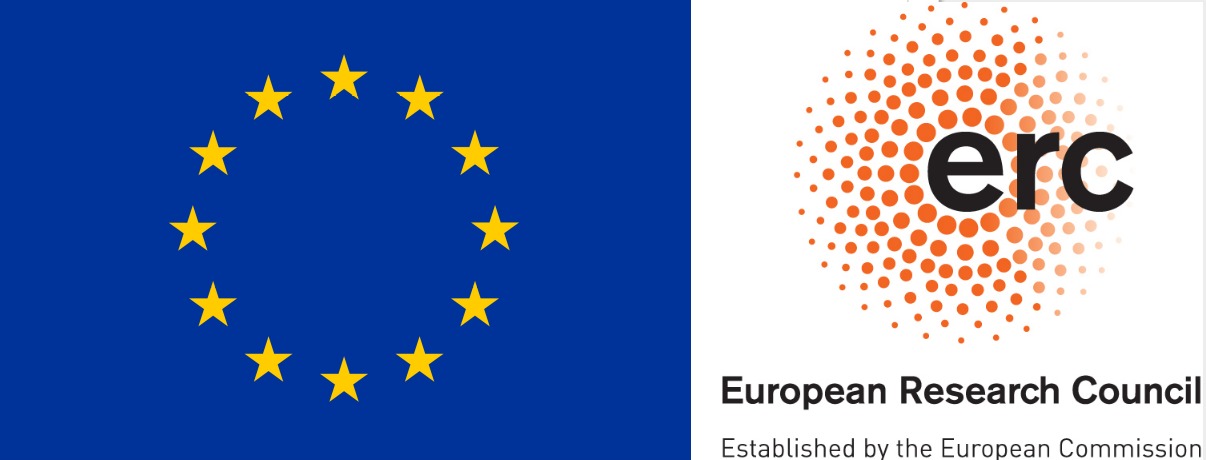} \\
        \vspace{4mm}
        \includegraphics[width=0.18\textwidth]{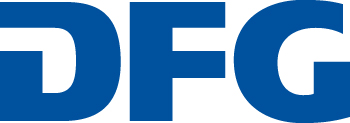}
    \end{center}
\vspace{-4mm}
\end{wrapfigure}
This work has received funding from the European Research Council (ERC)
(under the European Union's Horizon 2020 research and innovation
programme, grant agreement No 694537, project "SEQCLAS") and
the Deutsche Forschungsgemeinschaft (DFG; grant agreement NE 572/8-1, project "CoreTec").  
The GPU computing cluster was supported 
by DFG (Deutsche Forschungsgemeinschaft) under grant INST 222/1168-1 FUGG.
The work reflects only the authors' views and none of the funding agencies
is responsible for any use that may be made of the information it contains.

\bibliography{mybib}
\bibliographystyle{acl_natbib_nourl}

\end{document}